\documentclass[conference,a4paper]{IEEEtran}
\usepackage[utf8]{inputenc}
\usepackage[T1]{fontenc}
\usepackage{url}			 
\usepackage{cite}            

\usepackage[cmex10]{amsmath}  

\usepackage{mleftright}       

\usepackage{cgitIEEE}  
\mleftright

\usepackage{graphicx}         
\usepackage{booktabs}         

\usepackage{algorithmicx}     

\usepackage[caption=false,font=footnotesize]{subfig}

\usepackage{xcolor}
\usepackage{doi}
\usepackage{hyperref}
\hypersetup{ 
    colorlinks=true, 
    linkcolor=black, 
    citecolor=black, 
    urlcolor=black, 
    pdftitle={Your Title}, 
    pdfauthor={Your Name}, 
    bookmarksnumbered=true,
    bookmarksopen=true,
}

\IEEEoverridecommandlockouts

\begin{document}

\title{Equivalence of Empirical Risk Minimization to Regularization on the Family of $f$-Divergences
\thanks{This work is supported by the University of Sheffield ACSE PGR scholarships, the Inria Exploratory Action -- Information and Decision Making (AEx IDEM), H2020 RISE Project TESTBED2 under EU Grant 872172, and in part by a grant from the C3.ai Digital Transformation Institute.}
}


%
%
%
 \author{%
   \IEEEauthorblockN{Francisco Daunas\IEEEauthorrefmark{1}\IEEEauthorrefmark{2},
                     I\~naki Esnaola\IEEEauthorrefmark{1}\IEEEauthorrefmark{3},
                     Samir M.~Perlaza\IEEEauthorrefmark{2}\IEEEauthorrefmark{3}\IEEEauthorrefmark{4},
                     and H.~Vincent Poor\IEEEauthorrefmark{3}}
   \IEEEauthorblockA{\IEEEauthorrefmark{1}%
                     Dept. of ACSE, University of Sheffield,
                     Sheffield, United Kingdom.
                     \{jdaunastorres1, esnaola\}@sheffield.ac.uk}
   \IEEEauthorblockA{\IEEEauthorrefmark{2}%
                     INRIA,
                     Centre Inria d'Universit\'e C\^ote d'Azur,
                     Sophia Antipolis, France.
                     samir.perlaza@inria.fr}
   \IEEEauthorblockA{\IEEEauthorrefmark{3}%
                     Dept. of ECE, Princeton University, Princeton,
                     08544 NJ, USA.
                     poor@princeton.edu}
   \IEEEauthorblockA{\IEEEauthorrefmark{4}%
                     GAATI, Universit\'e de la Polyn\'esie Fran\c{c}aise,
                     Faaa, French Polynesia.}
 }

\maketitle

\begin{abstract}
  %
  The solution to empirical risk minimization with \mbox{$f$-divergence} regularization (ERM-$f$DR) is presented under mild conditions on $f$.
  Under such conditions, the optimal measure is shown to be unique.
  Examples of the solution for particular choices of the function $f$ are presented. Previously known solutions to common regularization choices are obtained by leveraging the flexibility of the family of $f$-divergences.
  These include the unique solutions to empirical risk minimization with relative entropy regularization (Type-I and Type-II).
  The analysis of the solution unveils the following properties of \mbox{$f$-divergences} when used in the ERM-$f$DR problem: $i\bigl)$ \mbox{$f$-divergence} regularization forces the support of the solution to coincide with the support of the reference measure, which introduces a strong inductive bias that dominates the evidence provided by the training data; and $ii\bigl)$ any \mbox{$f$-divergence} regularization is equivalent to a different \mbox{$f$-divergence} regularization with an appropriate transformation of the empirical risk function.
\end{abstract}

%
%
\section{Introduction}
\label{sec:introduction}

Empirical Risk Minimization (ERM) is a fundamental principle in machine learning. It is a tool for selecting a model from a given set by minimizing the empirical risk, which is the average loss or error induced by such a model on each of the labeled patterns available in the training dataset  
\cite{vapnik1964perceptron,vapnik1992principles}. In a nutshell, ERM aims to find a model that performs well on a given training dataset.
 However, ERM is prone to overfitting \cite{krzyzak1996nonparametric, deng2009regularized, arpit2017Memorization}, which affects the generalization capability of the selected model  \cite{tikhonov1993solution, horel1962application, bishop2006bookPattern}. To remediate this phenomenon, the solution of ERM must exhibit a small sensitivity to variations in the training dataset, which is often obtained via regularization \cite{bousquet2002stability, vapnik2015uniform, aminian2021exact, perlaza2024ERMRER, zou2024WorstCase}.  

In statistical learning theory, the ERM problem amounts to the minimization of the expected empirical risk over a subset of all probability measures that can be defined on the set of models. In this case,  regularization is often obtained by adding to the expected empirical risk a \textit{statistical distance} from the optimization measure, weighted by a regularization factor. Such a statistical distance is essentially a non-negative measure of dissimilarity between the optimization measure and the reference measure, which might be a $\sigma$-finite measure and not necessarily a probability measure, as shown in \cite{perlaza2024ERMRER} and \cite{Perlaza-ISIT-2022}.
A key observation is that the reference measure often represents prior knowledge or the inductive bias on the solution.

The notion of $f$-divergence, introduced in \cite{renyi1961measures}, and further studied in \cite{sason2016fdivergence} and \cite{csiszar1967information}, describes a family of hallmark statistical distances. A popular \mbox{$f$-divergence} is the relative entropy \cite{kullback1951information}, which due to its asymmetry, leads to two different problem formulations known as Type-I and Type-II ERM with relative entropy regularization (ERM-RER) \cite{InriaRR9508, InriaRR9474,Perlaza-ISIT2023a}. Relative entropy regularization also plays a central role in obtaining the worst-case data-generating probability measure introduced in\cite{zou2024WorstCase} and \cite{zou2024generalization}.    
The Type-I ERM-RER problem exhibits a unique solution, which is a Gibbs probability measure, independently of whether the reference measure is a probability measure or a $\sigma$-finite measure, as shown in \cite{perlaza2024ERMRER}.
The Type-II ERM-RER problem also has a unique solution when the reference measure is a probability measure. This solution exhibits properties that are analogous to those of the Gibbs probability measure \cite{InriaRR9474}.
Type-I ERM-RER appears to be the more popular regularized ERM problem, despite the fact that both types of regularization have distinct advantages. See for instance, \cite{robert2007bayesian, mcallester1998pacBayesian, valiant1984theory, shawe1997pac, cullina2018pac, vapnik1999overview, raginsky2016information, russo2019much, zou2009generalization, InriaRR9474, Perlaza-ISIT-2022, futami2023informationtheoretic} and references therein.

Optimization problems with \mbox{$f$-divergence} regularization have been explored before in \cite{teboulle1992entropic} and \cite{beck2003mirror} for the discrete case.
In \cite{alquier2021non}, the problem of non-exponentially weighted aggregation is studied. Such a problem involves an ERM with \mbox{$f$-divergence} regularization (ERM-$f$DR) identical to the one studied in this work. Nonetheless, the ERM-$f$DR imposes strong solution existence conditions on the function $f$, and thus, it holds for a reduced number of $f$-divergences.
This work presents the solution to the ERM-$f$DR problem using a method of proof that differs from those in  \cite{teboulle1992entropic, beck2003mirror} and \cite{alquier2021non} and goes along the lines of the methods in \cite{perlaza2024ERMRER, zou2024WorstCase} and \cite{InriaRR9474}, which rely on the notion of the Gateaux derivative \cite{gateaux1913fonctionnelles} and vector space methods \cite{luenberger1997bookOptimization}.

The method of proof favored in this paper enables the derivation of new results that have not been reported before. Firstly, the permissible values of the regularization factor that guarantee the existence of a solution are analytically characterized. 
Secondly, the obtained solution holds for a family of \mbox{$f$-divergences} that is larger than the one in \cite{alquier2021non}. For instance, the Type-II ERM-RER studied in \cite{Perlaza-ISIT2023a} and the ERM with Jensen-Shannon divergence regularization are both special cases of the ERM-$f$DR problem studied in this paper. These are examples of ERM-$f$DR problems that are not considered in \cite{alquier2021non}.
More importantly, the new method of proof allows showing that any \mbox{$f$-divergence} regularization is equivalent to a different \mbox{$f$-divergence} regularization with an appropriate transformation of the empirical risk function.

%
%
\section{Empirical Risk Minimization Problem}
\label{sec:ERMproblem}
Let~$\set{M}$,~$\set{X}$ and~$\set{Y}$, with~$\set{M} \subseteq \reals^{d}$ and~$d \in \ints$, be sets of \emph{models}, \emph{patterns}, and \emph{labels}, respectively.

A pair $(x,y) \in \mathcal{X} \times \mathcal{Y}$ is referred to as a \emph{labeled pattern} or \emph{data point},
%
and a \emph{dataset} is represented by the tuple $((x_1, y_1), (x_2, y_2), \ldots,(x_n, y_n))\in ( \set{X} \times \set{Y} )^n$.

Let the function~$h: \set{M} \times \mathcal{X} \rightarrow \mathcal{Y}$ be such that the label assigned to a pattern $x \in \set{X}$ according to the model $\thetav \in \set{M}$ is $h(\thetav,x)$.
Then, given a dataset
\begin{equation}
\label{EqTheDataSet}
\vect{z} = \big((x_1, y_1), (x_2, y_2 ), \ldots, (x_n, y_n )\big)  \in ( \set{X} \times \set{Y} )^n,
\end{equation}
the objective is to obtain a model $\thetav \in \set{M}$, such that, for all $i \inCountK{n}$, the label assigned to pattern $x_i$, which is $h(\thetav,x_i)$, is ``close'' to the label $y_i$.
This notion of ``closeness'' is formalized by the function
\begin{equation}
\label{EqRiskFunDef}
    \ell: \set{Y} \times \set{Y} \rightarrow [0, +\infty),
\end{equation}
such that the loss or risk induced by choosing the model $\thetav \in \set{M}$  with respect to the labeled pattern $(x_i, y_i)$, with $i\inCountK{n}$, is $\ell(h(\thetav,x_i),y_i)$.
The risk function $\ell$ is assumed to be nonnegative and to satisfy $\ell( y, y ) = 0$, for all $y\in\set{Y}$.

The \emph{empirical risk} induced by a model $\vect{\theta}$ with respect to the dataset $\vect{z}$ in~\eqref{EqTheDataSet} is determined by the function $\mathsf{L}_{\vect{z}}\!:\! \set{M} \rightarrow [0, +\infty)$, which satisfies
\begin{IEEEeqnarray}{rcl}
\label{EqLxy}
\mathsf{L}_{\vect{z}} (\vect{\theta} )  & \triangleq &
\frac{1}{n}\sum_{i=1}^{n}  \ell ( h(\vect{\theta}, x_i), y_i ).
\end{IEEEeqnarray}
The ERM problem with respect to the dataset $\vect{z}$ in~\eqref{EqTheDataSet} consists of the optimization problem:
\begin{equation}
\label{EqOPfunction}
\min_{\vect{\theta} \in \set{M}} \mathsf{L}_{\vect{z}}  (\vect{\theta}  ).
\end{equation}
The set of solutions to such a problem is denoted by
\begin{equation}
\label{EqHatTheta}
\set{T} ( \vect{z}  ) \triangleq \arg\min_{\vect{\theta} \in \set{M}}    \mathsf{L}_{\vect{z}}  (\vect{\theta}  ).
\end{equation}
Note that if the set $\set{M}$ is finite, the ERM problem in~\eqref{EqOPfunction} has a solution, and therefore, it holds that $\abs{\set{T}(\dset{z})}>0$.
Nevertheless, in general, the ERM problem does not always have a solution.
That is, there exist choices of the loss function $\ell$ and the dataset $\dset{z}$ that yield $\abs{\set{T}(\dset{z})}=0$.

%
%
\section{The ERM with $f$-Divergence Regularization}
\label{sec:ERMfDR}
\subsection{Preliminaries}
For ease of notation, the expected empirical risk with respect to a given measure is expressed via the following functional~$\foo{R}_{\dset{z}}$, defined below.
\begin{definition}[Expected Empirical Risk]
\label{DefEmpiricalRisk}
The expectation of the empirical risk $\mathsf{L}_{\vect{z}} (\vect{\theta} )$ in~\eqref{EqLxy}, when $\vect{\theta}$ is sampled from a probability measure $P \in \bigtriangleup(\set{M})$, is determined by the functional $\mathsf{R}_{\dset{z}}: \bigtriangleup(\set{M}) \rightarrow  [0, +\infty)$, such that
\begin{equation}
\label{EqRxy}
\foo{R}_{\dset{z}}( P ) = \int \foo{L}_{ \dset{z} } ( \thetav )  \diff P(\thetav).
\end{equation}
\end{definition}

In the following, the family of \mbox{$f$-divergences} is defined.
\begin{definition}[$f$-divergence \cite{csiszar1967information}]
\label{Def_fDivergence}
Let $f:[0,\infty)\rightarrow \reals$ be a convex function with $f(1)=0$ and $f(0) \triangleq \lim_{x\rightarrow 0^+}f(x)$.
Let $P$ and $Q$ be two probability measures on the same measurable space, with $P$ absolutely continuous with $Q$.
The \mbox{$f$-divergence} of $P$ with respect to $Q$, denoted by $\KLf{P}{Q}$,  is
\begin{equation}
\label{EqD_f}
\KLf{P}{Q} \triangleq \int f(\frac{\diff P}{\diff Q}(\thetav))  \diff Q(\thetav),
\end{equation}
where the function $\frac{\diff P}{\diff Q}$ is the Radon-Nikodym derivative of $P$ with respect to $Q$.
\end{definition}

In the case in which the function $f$ is continuous and differentiable, denote by $\dot{f}: [0, +\infty) \to \reals$ and $\dot{f}^{-1}: \reals \to [0, + \infty)$,
the derivative of~$f$ and the inverse of the function~$\dot{f}$, respectively.

The notation $\bigtriangleup(\set{M})$ denotes the set of all probability measures that can be defined upon the measurable space $\left(\set{M}, \BorSigma{\set{M}} \right)$, with $\BorSigma{\set{M}}$ being the Borel $\sigma$-field on $\set{M}$.
%
Given a probability measure $Q \in \bigtriangleup(\set{M})$ the set containing exclusively the probability measures in $\bigtriangleup(\set{M})$  that are absolutely continuous with respect to $Q$ is denoted by $\bigtriangleup_{Q}(\set{M})$. That is,
\begin{IEEEeqnarray}{rCl}
\label{DefSetTriangUp}
\bigtriangleup_{Q}(\set{M}) & \triangleq & \{P\in \bigtriangleup(\set{M}): P \ll Q \},
\end{IEEEeqnarray}
where the notation $P \ll Q$ stands for the measure $P$ being absolutely continuous with respect to the measure $Q$.

\subsection{Problem Formulation}

The ERM-$f$DR problem is parametrized by a probability measure $Q \in \bigtriangleup(\set{M})$, a positive real $\lambda$, and an \mbox{$f$-divergence} (Definition~\ref{Def_fDivergence}).
The measure $Q$ is referred to as the \emph{reference measure}, $\lambda$ as the \emph{regularization factor}, and $f$ as the \emph{regularization function}.
Given the dataset~$\dset{z} \in (\set{X} \times \set{Y})^n$ in~\eqref{EqTheDataSet}, the ERM-$f$DR problem, with parameters~$Q$ ,~$\lambda$ and $f$, consists of the following optimization problem:
\begin{IEEEeqnarray}{rcl}
\label{EqOp_f_ERMRERNormal}
    \min_{P \in \bigtriangleup_{Q}(\set{M})} & \quad \foo{R}_{\dset{z}} ( P )  + \lambda \KLf{P}{Q}.
\end{IEEEeqnarray}
\subsection{Solution to the ERM-$f$DR}

The solution of the ERM-$f$DR problem in~\eqref{EqOp_f_ERMRERNormal} is presented in the following theorem under the assumption that the function $f$ is strictly convex.

\begin{theorem}
\label{Theo_f_ERMRadNik}
If the function $f$ in~\eqref{EqOp_f_ERMRERNormal} is strictly convex, differentiable and there exists a $\beta$ in
\begin{subequations}
\label{EqfKrescConstrainAll}
\begin{equation}
\label{EqDefSetB}
\set{B} = \left\lbrace t\in \reals: \forall \vect{\theta} \in \supp Q , 0 <  \dot{f}^{-1} \left( -\frac{t + \foo{L}_{\vect{z}}(\thetav)}{\lambda} \right) \right\rbrace,
\end{equation}
such that
\begin{IEEEeqnarray}{rcl}
\label{EqEqualToABigOne}
\int \dot{f}^{-1}(-\frac{\beta + \foo{L}_{\dset{z}}(\thetav)}{\lambda}) \diff Q(\thetav)=1,
\end{IEEEeqnarray}
\end{subequations}
then the solution to the optimization problem in~\eqref{EqOp_f_ERMRERNormal}, denoted by $\Pgibbs{P}{Q} \in \bigtriangleup_{Q}(\set{M})$, is unique, and for all $\thetav \in \supp Q$ satisfies
\begin{equation}
\label{EqGenpdffDv}
\frac{\diff \Pgibbs{P}{Q}}{\diff Q} ( \thetav ) =  \dot{f}^{-1}(-\frac{\beta + \foo{L}_{\dset{z}}(\thetav)}{\lambda}).
\end{equation}
\end{theorem}
\begin{IEEEproof}
The proof is presented in Section 3.3 in \cite{InriaRR9521} 
\end{IEEEproof}
Theorem~\ref{Theo_f_ERMRadNik} implies that the Radon-Nikodym derivative $\frac{\diff \Pgibbs{P}{Q}}{\diff Q}$ in \eqref{EqGenpdffDv} is strictly positive.
A consequence of this observation is the following corollary.
\begin{corollary}
\label{coro_mutuallyAbsCont}
The probability measures~$Q$ and~$\Pgibbs{P}{Q}$ in~\eqref{EqGenpdffDv} are mutually absolutely continuous.
\end{corollary}
%
Corollary~\ref{coro_mutuallyAbsCont} reveals that, as is also the case with Type-II regularization, the support of the reference measure $Q$ establishes an inductive bias that cannot be overcome, regardless of the \mbox{$f$-divergence} choice. That is, the support of the solution is the support of the reference measure.
In a nutshell, the use of any strictly convex $f$-divergence regularization inadvertently forces the solution to coincide with the support of the reference independently of the training data.
Remarkably, from \cite[Corollary 23.5.1]{rockafellar1970conjugate} the function $\dot{f}^{-1}$ is the derivative of the convex conjugate of~$f$.

%
%
\subsection{Examples }
\label{sec:CaseResult}

Under the assumptions in Theorem~\ref{Theo_f_ERMRadNik} and assuming that $\set{B}$ in~\eqref{EqDefSetB} is not empty, this section presents the solutions for typical choices of the function~$f$.

\subsubsection{Kullback-Leibler Divergence}
\label{SubSubKL}
Let the function $f:(0,+\infty) \rightarrow \reals$ be such that $f(x) =  x\log(x)$,
\begin{subequations}
\label{ExampleKL_f}
whose derivative satisfies
\begin{IEEEeqnarray}{rCl}
\dot{f}(x) & = &  1 + \log(x).
\label{Eq_f_KL_s2}
\end{IEEEeqnarray}
In this case, the resulting $f$-divergence $\KLf{P}{Q}$ is the relative entropy of $P$ with respect to $Q$.
From~\eqref{Eq_f_KL_s2} and Theorem~\ref{Theo_f_ERMRadNik}, it holds that for all $\vect{\theta} \in \supp Q$,
\begin{IEEEeqnarray}{rCl}
\label{Eq_f_KL_dPdQ}
\frac{\diff \Pgibbs{P}{Q}}{\diff Q}(\thetav) & = & \exp(- \frac{\beta + \lambda + \foo{L}_{\dset{z}}(\thetav) }{\lambda}) \\
& = & \frac{\exp(- \frac{1}{\lambda}\foo{L}_{\dset{z}}(\thetav))}{\int \exp(- \frac{1}{\lambda}\foo{L}_{\dset{z}}(\nuv)) \diff Q (
\vect{\nuv})}.
\end{IEEEeqnarray}
\end{subequations}
This result has been independently reported by several authors in  \cite{Perlaza-ISIT-2022,Perlaza-ISIT2023b,perlaza2024ERMRER, raginsky2016information,zou2009generalization}, and the references therein.

\subsubsection{Reverse Relative Entropy Divergence}
\label{SubSubReverse}
Let the function $f:(0,+\infty) \rightarrow \reals$ be such that $f(x) = -\log(x)$,
\begin{subequations}
\label{ExampleRKL_f}
whose derivative satisfies
\begin{IEEEeqnarray}{rCl}
\dot{f}(x) & = & -\frac{1}{x}.
\label{Eq_f_rKL_s2}
\end{IEEEeqnarray}
In this case, the resulting $f$-divergence $\KLf{P}{Q}$ is the relative entropy of $Q$ with respect to $P$.
From~\eqref{Eq_f_rKL_s2} and Theorem~\ref{Theo_f_ERMRadNik}, it holds that for all $\vect{\theta} \in \supp Q$,
\begin{IEEEeqnarray}{rCl}
\label{Eq_f_rKL_dPdQ}
\frac{\diff \Pgibbs{P}{Q}}{\diff Q}(\thetav) & = & \frac{\lambda}{\beta + \foo{L}_{\dset{z}}(\thetav)}.
\end{IEEEeqnarray}
\end{subequations}
This result has been reported in \cite{zou2024WorstCase} and  \cite{Perlaza-ISIT2023a}.

\subsubsection{Jeffrey's Divergence}
Let the function $f:(0,+\infty) \rightarrow \reals$ be such that $f(x) = x\log(x) - \log(x)$,
\begin{subequations}
whose derivative satisfies
\begin{IEEEeqnarray}{rCl}
\dot{f}(x) & = & \log(x) + 1  - x^{-1}
\label{Eq_f_Jeff_s2}
\end{IEEEeqnarray}
In this case, the resulting $f$-divergence $\KLf{P}{Q}$ is Jeffrey's divergence between $P$ and $Q$.
From~\eqref{Eq_f_Jeff_s2} and Theorem~\ref{Theo_f_ERMRadNik}, it holds that for all $\vect{\theta} \in \supp Q$,
%
\begin{equation}
\label{Eq_f_Jeff_dPdQ}
	\frac{\diff \Pgibbs{P}{Q}}{\diff Q}\!(\thetav)
 \!=\! \exp(\!W_0\!(\exp(\!\frac{\beta\! +\! \lambda\!+\!\foo{L}_{\dset{z}}\!(\thetav)}{\lambda})\!)  \frac{\beta\!+\!\lambda\! +\! \foo{L}_{\dset{z}}\!(\thetav)}{\lambda}\!)\!,
\end{equation}
\end{subequations}
where the function $W_0:[0,\infty)\rightarrow [0,\infty)$ is the Lambert function, which for a function $g: \reals \to \reals$ such that $g(x) = x\exp(x)$ satisfies $W_0(g(x)) = x$.

\subsubsection{Hellinger Divergence}

Let the function $f:(0,+\infty) \rightarrow \reals$ be such that $ f(x) = (1-\sqrt{x})^2$,
\begin{subequations}
%
whose derivative satisfies
\begin{IEEEeqnarray}{rCl}
\label{Eq_f_Hell_s2}
\dot{f}(x) & = & 1 - \frac{1}{\sqrt{x}}.
\end{IEEEeqnarray}
In this case, the resulting $f$-divergence $\KLf{P}{Q}$ is Hellinger's divergence of $P$ with respect to $Q$.
From~\eqref{Eq_f_Hell_s2} and Theorem~\ref{Theo_f_ERMRadNik}, it holds that for all $\vect{\theta} \in \supp Q$,
\begin{IEEEeqnarray}{rCl}
\label{Eq_f_Hell_dPdQ}
\frac{\diff \Pgibbs{P}{Q}}{\diff Q}(\thetav) & = & (\frac{\lambda}{\beta + \lambda +\foo{L}_{\dset{z}}(\thetav)})^2.
\end{IEEEeqnarray}
\end{subequations}

\subsubsection{Jensen-Shannon Divergence}

Let the function $f:(0,+\infty) \rightarrow \reals$ be such that $f(x) = x \log(\frac{2x}{x+1}) + \log(\frac{2}{x+1})$,
\begin{subequations}
%
whose derivative satisfies
\begin{IEEEeqnarray}{rCl}
\label{Eq_f_JS_s2}
\dot{f}(x) & = & \log(2x) - \log(x + 1).
\end{IEEEeqnarray}
From~\eqref{Eq_f_JS_s2} and Theorem~\ref{Theo_f_ERMRadNik}, it holds that for all $\vect{\theta} \in \supp Q$,
%
\begin{IEEEeqnarray}{rCl}
\label{Eq_f_JS_dPdQ}
\frac{\diff \Pgibbs{P}{Q}}{\diff Q}(\thetav)
& = & \frac{1}{2\exp(\frac{\beta + \foo{L}_{\dset{z}}(\thetav)}{\lambda}) - 1}.
\end{IEEEeqnarray}
\end{subequations}

\subsubsection{$\chi^2$ Divergence}
Let the function $f:(0,\infty) \rightarrow \reals$ be such that $f(x) = (x-1)^2$,
\begin{subequations}
whose derivative satisfies
\begin{IEEEeqnarray}{rCl}
\dot{f}(x) & = & 2(x-1).
\label{Eq_f_X2_s2}
\end{IEEEeqnarray}
In this case, the resulting $f$-divergence $\KLf{P}{Q}$ is the Pearson-divergence, also known as, the $\chi^{2}$-divergence.
From~\eqref{Eq_f_X2_s2} and Theorem~\ref{Theo_f_ERMRadNik}, it holds that for all $\vect{\theta} \in \supp Q$,
%
%
\begin{IEEEeqnarray}{rCl}
\label{Eq_f_X2_dPdQ}
\frac{\diff \Pgibbs{P}{Q}}{\diff Q}(\thetav) & = & \frac{2\lambda - \beta - \foo{L}_{\dset{z}}(\thetav)}{2\lambda}.
\end{IEEEeqnarray}
\end{subequations}
%

%
%
\section{Analysis of Regularization Factor}
\label{sec:analysisRegFact}
%
This section studies the set of regularization factors for which the optimization problem in~\eqref{EqOp_f_ERMRERNormal} admits a solution. The analysis of the set $\set{B}$ defined in~\eqref{EqDefSetB} leads to the introduction of the \emph{normalization function} described below.
%
Let the function
\begin{subequations}
\label{EqDefNormFunction}
\begin{equation}
\label{EqDefMapNormFunction}
N_{Q, \dset{z}}: \set{A} \rightarrow \set{B},
\end{equation}
where $\set{A} \subseteq (0, \infty)$ represents the set of permissible regularization factors, and $\set{B}$ is defined in~\eqref{EqDefSetB}. This function is defined such that for all $\gamma \in \set{A}$,
\begin{equation}
\label{EqType2Krescaling}
N_{Q, \dset{z}}(\gamma) = t,
\end{equation}
\end{subequations}
where $t$ satisfies
\begin{IEEEeqnarray}{rCl}
\label{EqType2KrescConstrainKbarDef}
\int \dot{f}^{-1}(-\frac{t + \foo{L}_{\dset{z}}(\thetav)}{\gamma}) \diff Q(\thetav) = 1,
\end{IEEEeqnarray}
and the function $\foo{L}_{\dset{z}}$ is defined in~\eqref{EqLxy}.
Combining \eqref{EqDefNormFunction} and \eqref{EqType2KrescConstrainKbarDef}, it follows that
\begin{IEEEeqnarray}{rCl}
\label{EqReasonNisNormFoo}
\int \dot{f}^{-1}\left(-\frac{N_{Q, \dset{z}}(\lambda) + \foo{L}_{\dset{z}}(\thetav)}{\lambda}\right) \diff Q(\thetav) = 1,
\end{IEEEeqnarray}
which justifies calling the function $N_{Q, \dset{z}}$ as the \emph{normalization function}.
%
Some of the properties of interest of the function $N_{Q, \dset{z}}$ in~\eqref{EqDefNormFunction} are characterized by
%
\begin{equation}
\label{EqDefTStar}
t^{\star}_{Q,\dset{z}} \triangleq \inf  \set{B},
\end{equation}
under the assumption that $\set{B}\not = \emptyset$.
The following lemma introduces relevant properties of the function $N_{Q, \dset{z}}$ in~\eqref{EqDefNormFunction}.
\begin{lemma}
\label{lemm_InfDevKfDR}
The function~$N_{Q, \dset{z}}$ in~\eqref{EqDefNormFunction} is strictly increasing  and continuous.
\end{lemma}
\begin{IEEEproof}
The proof is presented in Appendix A in \cite{InriaRR9521} 
\end{IEEEproof}
 Since the function $f$ is strictly convex, then $\dot{f}^{-1}$ is strictly increasing, which in conjunction with Lemma~\ref{lemm_InfDevKfDR}, connects the term $t^{\star}_{Q,\dset{z}}$ to the infimum of the set $\set{A}$ in \eqref{EqDefMapNormFunction}. More specifically,
\begin{IEEEeqnarray}{rCl}
\lambda^{\star}\triangleq\inf \set{A} = N_{Q, \dset{z}}(t^{\star}_{Q,\dset{z}}).
\end{IEEEeqnarray}

\begin{lemma}
\label{lemm_fDR_kset}
If the set $\set{B}$ in~\eqref{EqDefSetB} is not empty, then it satisfies
\begin{IEEEeqnarray}{rCCCl}
\label{Eq_Type2KConstrainOpen}
(t^{\star}_{Q,\dset{z}}, \infty)& \subseteq &\set{B} & \subseteq & [t^{\star}_{Q,\dset{z}}, \infty).
\end{IEEEeqnarray}
Moreover, the set $\set{B}$ is identical to $[t^{\star}_{Q,\dset{z}}, \infty)$ if and only if
\begin{IEEEeqnarray}{rCl}
\label{EqT2lowlimDeltaIsInfty}
\int \dot{f}^{-1}(- t^{\star}_{Q,\dset{z}} -\foo{L}_{\dset{z}}(\thetav))\diff Q(\thetav)
& < & \infty,
\end{IEEEeqnarray}
with $t^{\star}_{Q,\dset{z}}$ defined in~\eqref{EqDefTStar}.
\end{lemma}
\begin{IEEEproof}
%
The proof is presented in Appendix B in \cite{InriaRR9521} 
\end{IEEEproof}
%

For the case in which $\set{B}$ is closed from the left, Lemma~\ref{lemm_InfDevKfDR} and Lemma~\ref{lemm_fDR_kset} imply the existence of a minimum regularization factor $\lambda^{\star} > 0$, with $\lambda^{\star} \in \set{A}$.
As a result, the solution to the optimization problem in~\eqref{EqOp_f_ERMRERNormal} only exists for regularization factors $\lambda \geq \lambda^{\star}$.
%
%
%
%
For the case in which $\set{B}$ is open from the left, the following lemma shows sufficient conditions for observing that $\set{A} = (0,\infty)$.

\begin{lemma}
\label{lemm_fDR_No_minRegF_nneg}
If the function $\dot{f}^{-1}$ in \eqref{EqGenpdffDv} is nonnegative and $\set{B}$ is not empty, then $\set{B}$ in~\eqref{EqDefSetB} is identical to $(t^{\star}_{Q,\dset{z}},\infty)$ and $\set{A}$ in \eqref{EqDefMapNormFunction} is identical to $(0,\infty)$, with $t^{\star}_{Q,\dset{z}}$ defined in~\eqref{EqDefTStar}.
\end{lemma}
\begin{IEEEproof}
%
The proof is presented in Appendix C in \cite{InriaRR9521} 
\end{IEEEproof}

Under the assumptions of Lemma~\ref{lemm_fDR_No_minRegF_nneg}, the ERM-$f$DR optimization in~\eqref{EqOp_f_ERMRERNormal} exhibits a unique solution for all $\lambda \in (0 ,\infty)$. This is the case of the \emph{Kullback-Leibler Divergence}, \emph{Jeffrey's Divergence} and \emph{Hellinger Divergence}, which makes them easy to implement regularizers as the constraint for existence in~\eqref{EqDefSetB} is always satisfied.

In the case of divergences such as the \emph{Reverse Relative Entropy Divergence}, \emph{Jensen-Shannon Divergence} and \emph{$\chi^2$ Divergence}, the existence of a lower bound on the regularization factor is dependent on the parameters of the ERM-$f$DR optimization in~\eqref{EqOp_f_ERMRERNormal}, which complicates their implementation in practical settings. The following examples illustrate this dependence on the parameters by providing cases in which $\set{B}$ in~\eqref{Eq_Type2KConstrainOpen} is the open set $(t^{\star}_{Q, \dset{z}},\infty)$ and closed set $[t^{\star}_{Q, \dset{z}},\infty)$  for the \emph{Reverse Relative Entropy Divergence}.


\begin{example}
\label{Ex1_Type2}
	Consider the ERM-$f$DR problem in \eqref{EqOp_f_ERMRERNormal} for $f(x) = -\log(x)$ and assume that: 
	$(a)$ $\set{M} = \set{X} = \set{Y} = [0,\infty)$; $(b)$ $\dset{z} = (1,0)$ and $(c)$ $Q \ll \mu$, with $\mu$ the Lebesgue measure, such that for all $\thetav \in \supp Q$,
	\begin{subequations}
	\label{EqEx1dQdmu_all}
	\begin{IEEEeqnarray}{rCl}
	\label{EqEx1dQdmu}
	 \frac{\diff Q}{\diff \mu}(\thetav) & = &  4\thetav^2\exp(-2\thetav).
	\end{IEEEeqnarray} 
	Let also the function $h:\set{M}\times\set{X}\rightarrow\set{Y}$ be
	$h(\thetav,x) =  x\thetav$, 
	%
	%
	and the risk function $\ell$ in \eqref{EqRiskFunDef} be
	\begin{IEEEeqnarray}{rCl}
	\label{EqEx1ellDef}
		\ell(h(\thetav,x),y)& = & (x\thetav-y)^2,
	\end{IEEEeqnarray}
	which implies
	\begin{IEEEeqnarray}{rCl}
		\foo{L}_{\dset{z}}(\thetav)
		& = & (x\thetav-y)^2,
	\end{IEEEeqnarray}
	\end{subequations}
	with the function $\foo{L}_{\dset{z}}$ defined in \eqref{EqLxy}.
	Under the current assumptions, the objective of this example is to show that $\set{B} = [t^{\star}_{Q,\dset{z}},\infty)$. For this purpose, it is sufficient to show that the inequality in \eqref{EqT2lowlimDeltaIsInfty} holds.
	From Theorem~\ref{Theo_f_ERMRadNik}, it follows that $\Pgibbs{P}{Q}$ in \eqref{EqGenpdffDv} satisfies for all $\thetav \in \supp Q$,  
	\begin{IEEEeqnarray}{rCl}
	\label{EqRDdPdmu}
	 \frac{\diff \Pgibbs{P}{Q}}{\diff \mu}(\thetav) & = & \frac{\lambda}{\foo{L}_{\dset{z}}(\thetav)+\beta}4\thetav^2\exp(-2\thetav),
	\end{IEEEeqnarray}
	with $\beta$ satisfying \eqref{EqfKrescConstrainAll}.
	Thus,
	\begin{subequations}
	\label{EqEx1dPdmu}
	\begin{IEEEeqnarray}{rCl}
	\label{EqEx1dPdmu_s1}
	 \int \frac{1}{\foo{L}_{\dset{z}}(\thetav)+t^{\star}_{Q,\dset{z}}} \diff Q(\thetav)
	 & = & \int^{\infty}_{0} 4\exp(-2\thetav) \diff \thetav \label{EqEx1dPdmu_s6}\\
	 & = & 2,
	\end{IEEEeqnarray}
	\end{subequations}
	where equality \eqref{EqEx1dPdmu_s1} follows from \eqref{EqEx1dQdmu_all}, the assumption that $(x,y) = (1,0)$ and the fact that $t^{\star}_{Q, \dset{z}} = 0$.
	Finally, the function $N_{Q, \dset{z}}$ in \eqref{EqDefNormFunction} satisfies $N_{Q, \dset{z}}(\frac{1}{2})= 0$, which implies $t^{\star}_{Q,\dset{z}}\in \set{B}$, that is, $\set{B} = [0,\infty)$ and $\set{A} = [\frac{1}{2},\infty)$.
\end{example}

%

\begin{example}
\label{Ex3_Type2}
	Consider Example~\ref{Ex1_Type2} with \mbox{$\dset{z} = (1,1)$}. Under the current assumptions, the objective of this example is to show that $\set{B} = (t^{\star}_{Q, \dset{z}},\infty)$. For this purpose, it is sufficient to show that the inequality in \eqref{EqT2lowlimDeltaIsInfty} does not hold:
	\begin{subequations}
	\begin{IEEEeqnarray}{rCl}
	\label{EqProofT2Ex3Integral_s1}
	\int \frac{1}{\foo{L}_{\dset{z}}(\thetav)-t^{\star}_{Q,\dset{z}}} \diff Q(\thetav)
	 & = & \int^{\infty}_{0}\frac{4\thetav^2\exp(-2\thetav)}{(\thetav-1)^2} \diff \thetav \label{EqProofT2Ex3Integral_s4}\\
	 & = & \infty. \label{EqProofT2Ex3Integral_s5}
	\end{IEEEeqnarray}
	\end{subequations}
	where equality \eqref{EqProofT2Ex3Integral_s1} follows from equality \eqref{EqEx1dQdmu}; equality \eqref{EqProofT2Ex3Integral_s4} follows from the assumption that $(x,y) = (1,1)$ and the fact that $t^{\star}_{Q, \dset{z}} = 0$; and the equality \eqref{EqProofT2Ex3Integral_s5} follows from an algebraic development.
	%
	Finally,  the function $N_{Q, \dset{z}}$ in \eqref{EqDefNormFunction} is undefined at zero, which implies $t^{\star}_{Q,\dset{z}} \not\in \set{B}$, that is, $\set{B} = (0,\infty)$.
\end{example}

%

These examples illustrate that even if the reference measure $Q$ and functions $\ell$ and $h$ in \eqref{EqLxy} are fixed, the set $\set{B}$ might be either $[t^{\star}_{Q, \dset{z}}, \infty)$ or  $(t^{\star}_{Q, \dset{z}}, \infty)$ depending on the dataset $\dset{z}$.
This observation underscores that the existence of the minimum regularization factor $\lambda^\star$ is coupled on the specific choices of $Q$, $\ell$, $f$, and $\dset{z}$.
 

%
%
\section{Equivalence of the $f$-Regularization via Transformation of the Empirical Risk}
\label{sec:equivalence}
This section shows that given two strictly convex and differentiable functions $f$ and $g$ that satisfy the conditions in Definition~\ref{Def_fDivergence}, there exists a function $v:[0,\infty)\rightarrow \reals$, such that the solution to the optimization problem in~\eqref{EqOp_f_ERMRERNormal} is identical to the solution of the following problem:
\begin{IEEEeqnarray}{rcl}
\label{EqOpERMLinkT1_T2}
    \min_{P \in \bigtriangleup_{Q}(\set{M})} & \quad & \int v(\foo{L}_{\dset{z}}(\thetav)) \diff P(\thetav) + \lambda\Divf[g]{P}{Q},
\end{IEEEeqnarray}
with $\lambda$ and $Q$ in~\eqref{EqOp_f_ERMRERNormal}.
The main result of this section is presented in the following theorem.
%
\begin{theorem}
\label{Theo_ERMfDRType1To2}
Let $f$ and $g$ be two strictly convex and differentiable functions satisfying the conditions in Definition~\ref{Def_fDivergence}. If the problem in~\eqref{EqOp_f_ERMRERNormal} possesses a solution, then
\begin{IEEEeqnarray}{rCl}
\IEEEeqnarraymulticol{3}{l}{
\min_{P \in \bigtriangleup_{Q}(\set{M})}  \int \foo{L}_{\dset{z}}(\thetav) \diff P(\thetav)  + \lambda \Divf[f]{P}{Q}
}\nonumber \\ \quad \label{EqDefHfromF2G}
& = &  \min_{P \in \bigtriangleup_{Q}(\set{M})}  \int v(\foo{L}_{\dset{z}}(\thetav)) \diff P(\thetav)   + \lambda\Divf[g]{P}{Q},
\end{IEEEeqnarray}
where the function $v:[0,\infty) \rightarrow \reals$ is such that
\begin{IEEEeqnarray}{rCl}
\label{EqFromf2gfDR}
v(t)
& = & \lambda\dot{g}(\dot{f}^{-1}(-\frac{N_{Q,\dset{z}}(\lambda)+t}{\lambda}))-N'_{Q,\dset{z}}(\lambda),
\end{IEEEeqnarray}
where $N_{Q,\dset{z}}$ and $N'_{Q,\dset{z}}$ are the normalization functions of the optimization problems in \eqref{EqOp_f_ERMRERNormal} and \eqref{EqOpERMLinkT1_T2}.
\end{theorem}
\begin{IEEEproof}
  Note that from Theorem~\ref{Theo_f_ERMRadNik} the functions $f$ and $g$ are differentiable and strictly convex. Hence, the functional inverse of the derivative is well-defined from the fact that $\dot{f}$ and $\dot{g}$ are strictly increasing and bijective.
  Denote by $\Pgibbs{\hat{P}}{Q}$ the solution to the optimization problem in~\eqref{EqOpERMLinkT1_T2}. Then, from Theorem~\ref{Theo_f_ERMRadNik}, for all $\thetav\in \supp{Q}$, it follows that
 \begin{subequations}
\label{EqProoT1ToT2}
\begin{IEEEeqnarray}{rcl}
\frac{\diff \Pgibbs{\hat{P}}{Q}}{\diff Q}(\thetav)
& = & \dot{g}^{-1}(-\frac{N'_{Q,\dset{z}}(\lambda) + v(\foo{L}_{\dset{z}}(\thetav))}{\lambda})
\label{EqProoT1ToT2_s1} \\
& = & \dot{g}^{-1}(\dot{g}(\dot{f}^{-1}(-\frac{N_{Q,\dset{z}}(\lambda) + \foo{L}_{\dset{z}}(\thetav)}{\lambda})))
\label{EqProoT1ToT2_s2}\qquad\\
& = & \dot{f}^{-1}(-\frac{N_{Q,\dset{z}}(\lambda) + \foo{L}_{\dset{z}}(\thetav)}{\lambda})
\label{EqProoT1ToT2_s3}\\
& = &  \frac{\diff \Pgibbs{P}{Q}}{\diff Q}(\thetav) \label{EqProoT1ToT2_s4},
\end{IEEEeqnarray}
\end{subequations}
where the equality in~\eqref{EqProoT1ToT2_s2} follows from~\eqref{EqFromf2gfDR}, which completes the proof.
\end{IEEEproof}

Theorem~\ref{Theo_ERMfDRType1To2} establishes an equivalence between two ERM problems subject to different $f$-divergence regularizations. Such equivalence can always be established as long as the corresponding divergences are defined by strictly convex and differentiable functions.
%
%
More importantly, for all strictly convex $f$ functions, the solution to the corresponding
ERM with $f$-divergence regularization is mutually absolutely continuity with respect to the reference measure.
%


The following example illustrates the equivalence between two $f$-divergence regularizations.
The objective of this example is to demonstrate the equivalence of the solutions to the optimization problems in~\eqref{EqOp_f_ERMRERNormal} and \eqref{EqOpERMLinkT1_T2}.
\begin{example}
\label{Ex4_Type2}
	Consider the optimization problems in \eqref{EqOp_f_ERMRERNormal} and~\eqref{EqOpERMLinkT1_T2} with $f(t) = -\log(t)$ and $g(t) = -\log(t)$, respectively.  
	The solution to the optimization problem in \eqref{EqOp_f_ERMRERNormal} is described in Section \ref{SubSubKL}.
	Denote by $\Pgibbs{\hat{P}}{Q}$ the solution to the optimization problem in \eqref{EqOpERMLinkT1_T2}.
	From Theorem \ref{Theo_f_ERMRadNik}, it follows that for all $\thetav \in \supp Q$,
	\begin{IEEEeqnarray}{rcl}
	\label{Eq_g_example4}
		\frac{\diff \Pgibbs{\hat{P}}{Q}}{\diff Q}(\thetav)
		& = & \frac{\lambda}{v(\foo{L}_{\dset{z}}(\thetav))+ \beta},
	\end{IEEEeqnarray}
	where the function $v$ is defined in \eqref{EqFromf2gfDR} and for the $f$ and $g$ of this example satisfies for all $\thetav \in \supp Q$,
	\begin{IEEEeqnarray}{rCl}
		v(\foo{L}_{\dset{z}}(\thetav)) & = &\! \lambda\exp(\!\frac{\foo{L}_{\dset{z}}(\thetav)}{\lambda} \! + \! \log(\int \! \exp(\!-\frac{\foo{L}_{\dset{z}}(\nuv)}{\lambda})\diff Q(\nuv)\!)\!)\nonumber\\
		&  & - \>\beta.
		\label{Eq_V_example4}
	\end{IEEEeqnarray}
	Plugging \eqref{Eq_V_example4} into \eqref{Eq_g_example4} yields
	\begin{IEEEeqnarray}{rcl}
		\frac{\diff \Pgibbs{\hat{P}}{Q}}{\diff Q}(\thetav) & = & \frac{\exp(- \frac{1}{\lambda}\foo{L}_{\dset{z}}(\thetav))}{\int \exp(- \frac{1}{\lambda}\foo{L}_{\dset{z}}(\nuv)) \diff Q (\vect{\nuv})},
	\end{IEEEeqnarray}
	which is the solution to the optimization problem in~\eqref{EqOp_f_ERMRERNormal} presented in Section \ref{SubSubKL}.
\end{example}
\section{Conclusions}\label{SecConclusions}

This work has presented the solution to the ERM-$f$DR problem under mild conditions on $f$, namely, $(a)$ strict convexity; and $(b)$ differentiability.
Under these conditions, the optimal measure is shown to be unique and sufficient conditions for the existence of the solution are presented.
%
%
This result unveils the fact that all parameters are involved in guaranteeing the existence of a solution.
%
Remarkably, $f$-divergence regularizers that satisfy the conditions above, can be transformed into a different $f$-divergence regularizer by a transformation of the empirical risk. The mutual absolute continuity of the ERM-$f$DR solutions to the reference measure can be understood in light of the equivalence between the regularization.
The analytical results have also enabled us to provide insights into choices of \mbox{$f$-divergences} for algorithm design in statistical machine learning.


%
\IEEEtriggeratref{23}
\bibliographystyle{IEEEtran}
\bibliography{iEEEtranBibStyle.bib}
%

\newpage




\end{document}